\documentclass[journal,transmag,conference]{IEEEtran}
\IEEEoverridecommandlockouts
\usepackage{cite}
\usepackage{amsmath,amssymb,amsfonts}
\usepackage{algorithmic}
\usepackage{graphicx}
\usepackage{textcomp}
\usepackage{xcolor}
\usepackage{float}
\usepackage{tabularx}
\usepackage{caption}
\usepackage{subcaption}

\def\BibTeX{{\rm B\kern-.05em{\sc i\kern-.025em b}\kern-.08em
    T\kern-.1667em\lower.7ex\hbox{E}\kern-.125emX}}
\begin{document}

\title{ Multisensor Data Fusion for Reliable Obstacle Avoidance}

\author{\IEEEauthorblockN{Thanh Nguyen Canh\IEEEauthorrefmark{1},
Truong Son Nguyen\IEEEauthorrefmark{1},
Cong Hoang Quach\IEEEauthorrefmark{1},
Xiem HoangVan\IEEEauthorrefmark{1} and
\\ Manh Duong Phung\IEEEauthorrefmark{2}}
\IEEEauthorblockA{\IEEEauthorrefmark{1}University of Engineering and Technology, Vietnam National University, Hanoi, Vietnam}
\IEEEauthorblockA{\IEEEauthorrefmark{2}Fulbright University Vietnam, Ho Chi Minh City, Vietnam}}

\maketitle

\begin{abstract}
Abstract---In this work, we propose a new approach that combines data from multiple sensors for reliable obstacle avoidance. The sensors include two depth cameras and a LiDAR arranged so that they can capture the whole 3D area in front of the robot and a 2D slide around it. To fuse the data from these sensors, we first use an external camera as a reference to combine data from two depth cameras. A projection technique is then introduced to convert the 3D point cloud data of the cameras to its 2D correspondence. An obstacle avoidance algorithm is then developed based on the dynamic window approach. A number of experiments have been conducted to evaluate our proposed approach. The results show that the robot can effectively avoid static and dynamic obstacles of different shapes and sizes in different environments. 
\end{abstract}


\section{Introduction}
Obstacle avoidance is essential for the safe operation of mobile robots. This task involves the design of a sensory system for data collection and algorithms for data processing and motion planning. The sensory system may consist of single or multiple sensors depending on the working environment and application \cite{Cadena2016}. In single-sensor methods, cameras and LiDAR are widely used due to their capability to collect rich information about the environment \cite{Davison2003,8325781,PHUNG201725}. Cameras can collect color information which allows for extracting features of the environment and obstacles. This information however is sensitive to lighting conditions causing difficulties in creating stable detection algorithms. Recent cameras can collect both color and depth information to better address the lighting issue. However, its resolution and field of view are still low. LiDAR sensors, on the other hand, can produce precise distance measurements at high frequencies. It therefore can achieve a low drift motion estimation with an acceptable computational complexity when applied to the obstacle avoidance  problem \cite{Debeunne2020}. However, 2D LiDARs only carries 2D information making it insufficient to cover the whole environment. 3D LiDARs are available but their price is still high for most applications. Therefore, a more relevant solution is to use multiple sensors to form a more complete picture of the environment for obstacle avoidance. In \cite{kumar2010sensor,5940576}, laser and vision sensors are used for depth estimation and obstacle avoidance. In \cite{6466599}, a compass sensor, a laser range finder, and an omni-directional camera are fused within a Kalman filter for location estimation and navigation. Other sensors such as Inertial Measurement Units (IMU) \cite{Huang2019}, Ultra-Wide Band (UWB) \cite{Liu2011}, Wifi \cite{He2016}, and Bluetooth \cite{Faragher2015} can also be used for navigation and obstacle avoidance.  One of the main issues with the use of multiple sensors is the need for calibration among the sensors. Some works address this by finding the corresponding points or edges among different sensors \cite{GONG2013394, Park2014,GarcaMoreno2013}. Some other works try to estimate the transformation matrices between the sensors \cite{Vasconcelos2012, Li2015}. However, the choice of sensors and methods to calibrate them is still a challenging problem due to their differences in the information acquired, sampling frequency, accuracy, resolution, bandwidth, etc.

Apart from sensors, data processing algorithms are also important for obstacle avoidance. A number of algorithms have been developed for this task such as the vector field histogram (VFH) \cite{Borenstein1991}, dynamic-window approach \cite{Fox1997}, occupancy grid algorithm \cite{Elfes1989}, and artificial potential field method \cite{Yunfeng2000}. Their effectiveness however varies depending on the sensors used and the environments in that the robot operates. Obstacle avoidance algorithms are therefore should be properly selected and optimized for the sensors and environments used.

In this paper, we propose a system that includes several sensors and a control algorithm for obstacle avoidance. The sensors include two depth cameras placed on two sides of the robot and a LiDAR placed in the middle of the cameras. The control algorithm is based on the dynamic-window approach, but uses the data fused from the sensors to calculate the input velocities. Our contributions in this study include:
\begin{itemize}
\item A multi-sensor obstacle avoidance system that combines both cameras and LiDAR to obtain the ability to avoid challenging objects such as tables, chairs, and moving people.
\item A method to fuse data from the cameras and LiDAR to create a multi-layer map for navigation and obstacle avoidance.
\item A reliable obstacle avoidance algorithm that can be implemented in an embedded computer to run in real time.
\end{itemize}

\section{Sensor data fusion}
\label{section:fusion}
This section describes our robot platform and algorithms developed for data fusion.

\begin{figure}[!ht]
    \centering
    \includegraphics[width=0.7\linewidth]{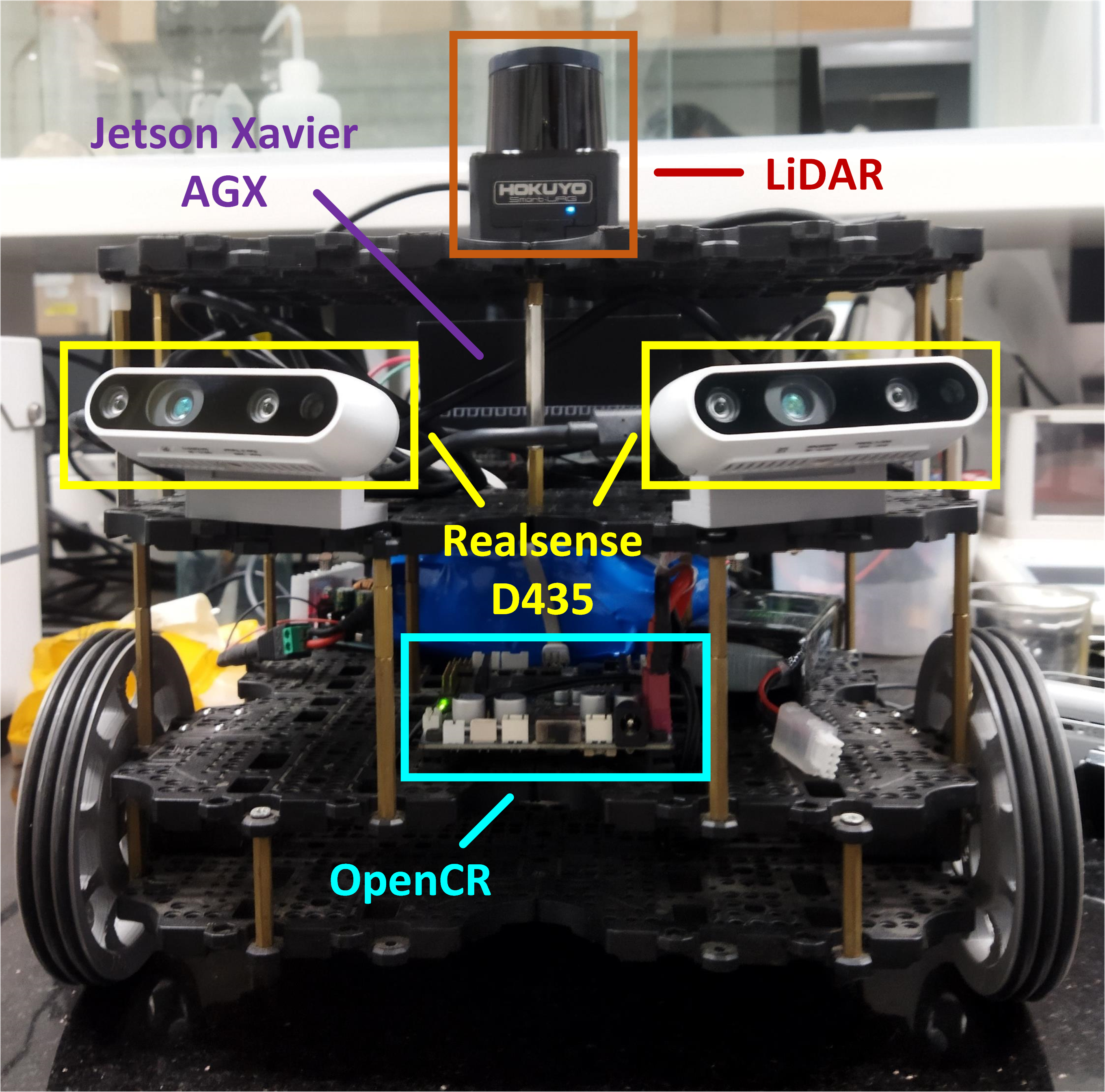}
    \caption{The robot platform}
    \label{fig:robotplatform}
\end{figure}

\subsection{Robot platform}
The robot used in this study is a differential drive mobile robot equipped with a sensory system including a 2D LiDAR and two depth cameras as shown Fig. \ref{fig:robotplatform}. The LiDAR is a Hokuyo URG-20LX. The cameras are Intel Realsense D435i. The robot is driven by two Dynamixel servo motors and controlled by two onboard computers. The first computer is an OpenCR board with a built-in inertial measurement unit (IMU) for low-level control. The second one is an embedded computer called Jetson Xavier AGX used for high-level data processing and navigation. The middleware used is ROS Melodic. 

\begin{figure}[!ht]
    \centering
    \includegraphics[width=0.65\linewidth]{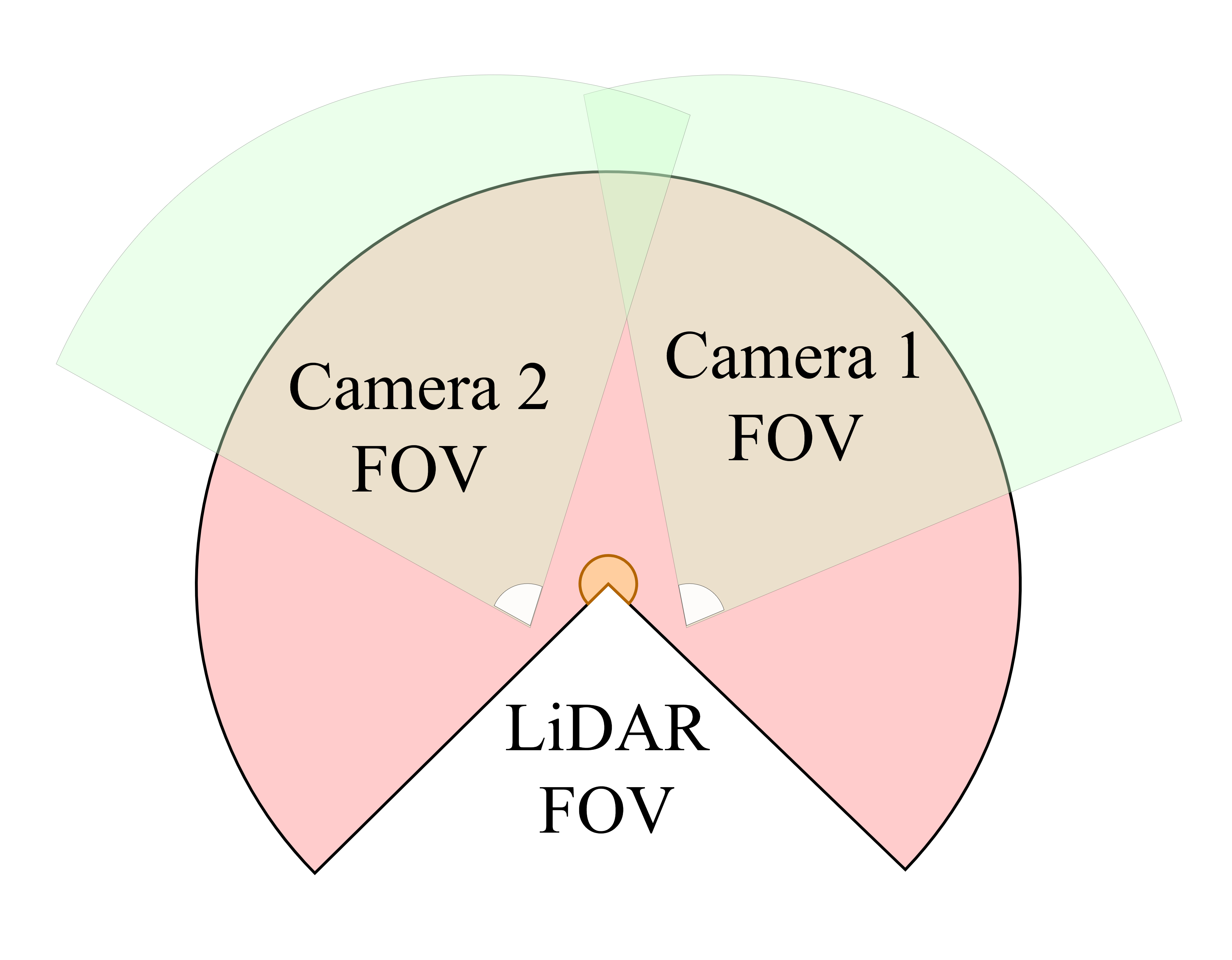}
    \caption{Field of views of the LiDAR and cameras system}
    \label{fig:FOV}
\end{figure}

For reliable navigation and obstacle avoidance, we placed the cameras at a tilted angle of 15 degrees upwards. Each camera has a horizontal field of view (FOV) of 87 degrees. The LiDAR is placed between the two cameras and has a FOV of 240 degrees. Figure \ref{fig:FOV} illustrates the overall FOV of the sensor system. With this arrangement, the LiDAR can identify objects around the robots and collect depth information about the environment for localization and mapping. The cameras, on the other hand, collect information on the whole 3D area in front of the robot and when combined with LiDAR data will provide sufficient information for obstacle avoidance. The data fusion is carried out in two stages, between the cameras and between the LiDAR and the cameras. 

\subsection{Depth camera data fusion}
\begin{figure}[h]
    \centering
    \includegraphics[width=0.45\textwidth]{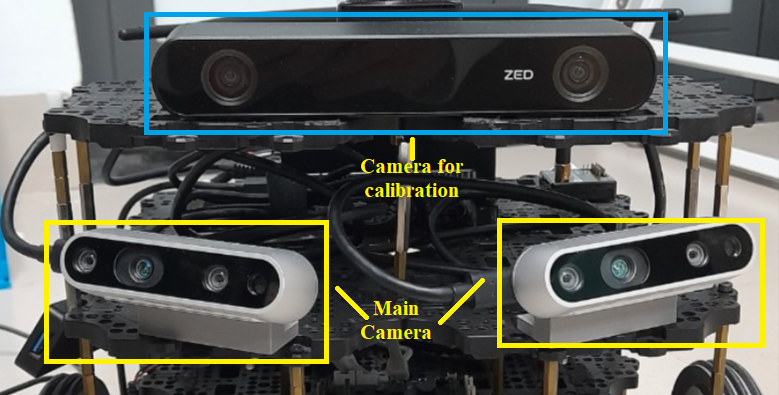}
    \caption{Two Realsense D435 depth cameras for obstacle avoidance and one Zed2 external camera for calibration}
    \label{fig:calib}
\end{figure}

Fusing data from two cameras is the problem of finding the transformation matrix between those cameras' coordinates. Although our two cameras have a small overlap region, the information in this region is insufficient to use for calibration. Hence, we propose to use an external camera as a reference to calculate the transformation matrix, as shown in Fig. \ref{fig:calib}. We respectively estimate the transformation matrices between this external camera and the first and second cameras, and then combine those matrices to obtain the overall transformation matrix.

To obtain each transformation matrix, an ArUco marker is used to convert the camera's coordinate to the coordinate attached to the ArUco marker. This marker is composed of a black border and an inner binary matrix that encodes its identifier \cite{ArUco}. By detecting the marker and its identifier, points on the marker can be mapped to their correspondence in the image plane. The Perspective-n-Point (PnP) problem then can be formulated and solved using methods such as EPnP \cite{lepetit2009epnp} to find the camera pose. Denote ${ }^{c} \mathbf{M}_{o}$ as the matrix representing the camera pose in the marker's coordinate frame. A point $( X_{c}, Y_{c}, Z_{c} )$ in the camera's coordinate frame then can be represented by $ ( X_{o}, Y_{o}, Z_{o} ) $ in the marker's coordinate frame as follows:

\begin{equation}
\begin{aligned}
\begin{bmatrix}
X_{c} \\
Y_{c} \\
Z_{c} \\
1
\end{bmatrix} &={ }^{c} \mathbf{M}_{o}
\begin{bmatrix}
X_{o} \\
Y_{o} \\
Z_{o} \\
1
\end{bmatrix} 
=\begin{bmatrix}
{ }^{c} \mathbf{R}_{o} & { }^{c} \mathbf{T}_{o} \\
0_{1 \times 3} & 1
\end{bmatrix}
\begin{bmatrix}
X_{o} \\
Y_{o} \\
Z_{o} \\
1
\end{bmatrix},
\end{aligned}
\end{equation}
where ${ }^{c} \mathbf{R}_{o} $ is the rotation matrix between the maker's coordinate frame and the camera's coordinate frame, and ${ }^{c} \mathbf{T}_{o}$ is the translation between the two coordinate frames. 

Denote ${ }^{c_{1}} \mathbf{M}_{o}$ and ${ }^{c_{2}} \mathbf{M}_{o}$ as the matrices representing the poses of camera 1 and 2, respectively. The transformation between the two cameras is then computed as:

\begin{equation}
\begin{aligned}
{ }^{c_{2}} \mathbf{M}_{c_{1}} = { }^{c_{2}} \mathbf{M}_{o} \times { }^{o} \mathbf{M}_{c_{1}} 
={ }^{c_{2}} \mathbf{M}_{o} \times \left({ }^{c_{1}} \mathbf{M}_{o}\right)^{-1} \\
=\left[\begin{array}{cc}
{}^{c_2} \mathbf{R}_{o} & { }^{c_{2}} \mathbf{T}_{o} \\
0_{3 \times 1} & 1
\end{array}\right] \times \left[\begin{array}{cc}
{}^{c_1} \mathbf{R}_{o}^{T} & -{ }^{c_{1}} \mathbf{R}_{o}^{T} \cdot{ }^{c_{1}} \mathbf{T}_{o} \\
0_{1 \times 3} & 1
\end{array}\right].
\end{aligned}
\end{equation}

\subsection{LiDAR and depth camera data fusion}
\label{sec:fusionlidarcamera}
\begin{figure}[!ht]
    \centering
    \begin{subfigure}[b]{0.2\textwidth}
    \centering
    \includegraphics[width=\textwidth]{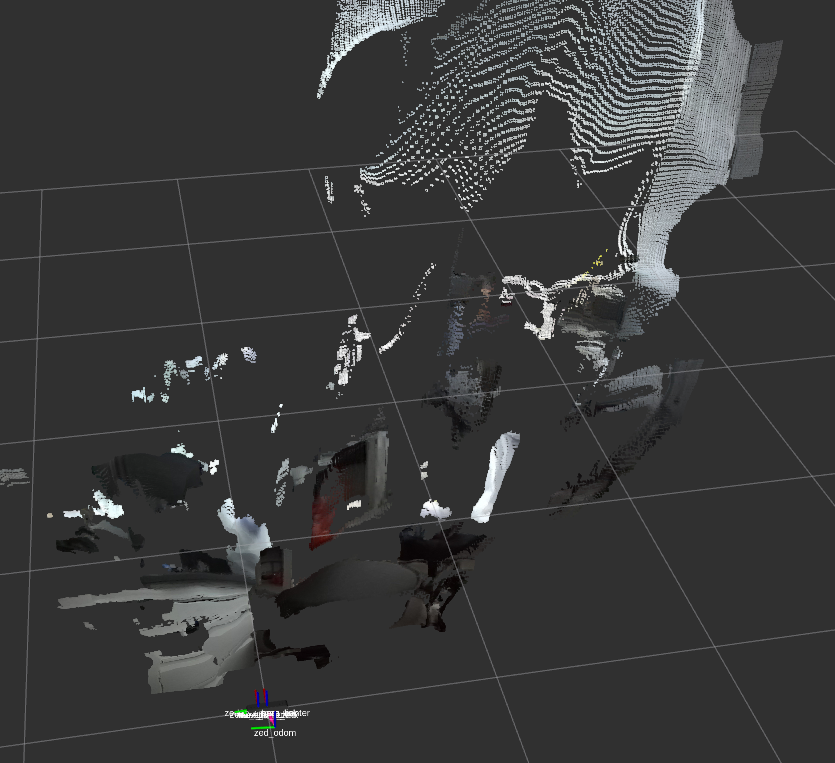}
    \caption{3D point cloud}
    \label{fig:3dpointcloud}
    \end{subfigure}
    \begin{subfigure}[b]{0.245\textwidth}
    \centering
    \includegraphics[width=\textwidth]{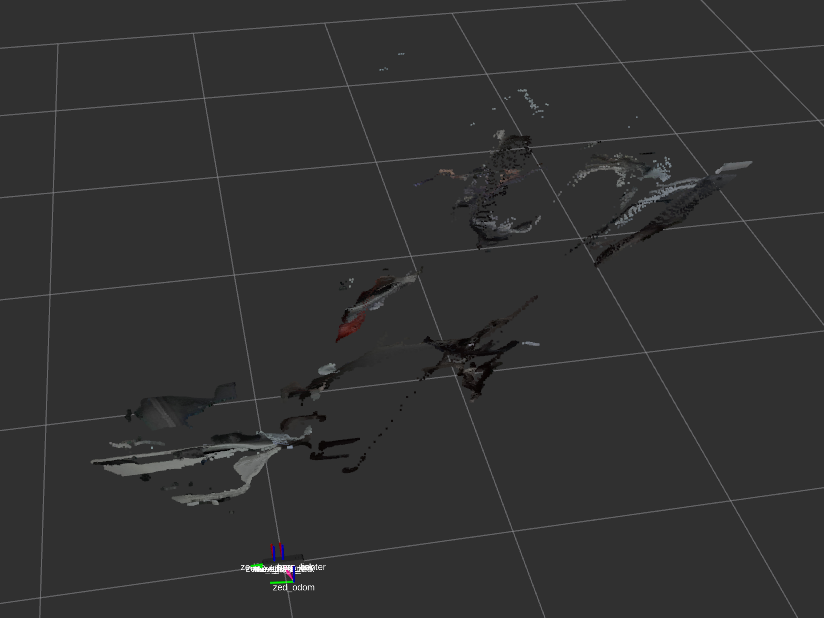}
    \caption{Projected point cloud}
    \label{fig:2Dpointcloud}
    \end{subfigure}
    
    \caption{Projection of a 3D point cloud to 2D}
    \label{fig:projectdata}
\end{figure}

Data fused from two cameras is a 3D point cloud as illustrated in Fig.\ref{fig:3dpointcloud}. To combine it with 2D data obtained from the LiDAR, it is necessary to convert the 3D point cloud to its 2D form. We do it by projecting the 3D point cloud to the scanning plane of the LiDAR. The projection is carried out by first calculating the coordinates of 3D points with respect to the frame attached to the LiDAR and then omitting the $z$ component.  

Since the camera is tilted upwards, we first rotate the frame attached to it to be in parallel with the LiDAR's plane. We then translate it to the origin of the LiDAR to align their two frames. Let $^LR_c$ and $^LT_c$ be respectively the rotation and translation matrices between the camera and the LiDAR. A point $P_c = (x_c,y_c,z_c)$ in the camera's frame is represented as $P_L = (x_L,y_L,z_L)$ in the LiDAR's frame as follows:

\begin{equation}
\label{eq:rot_point}
\begin{aligned}
\begin{bmatrix}
x_{L} \\
y_{L} \\
z_{L} \\
1 \\
\end{bmatrix} 
=\begin{bmatrix}
^LR_c & ^LT_c\\
0_{3 \times 1} & 1 \\
\end{bmatrix}
\begin{bmatrix}
x_{c} \\
y_{c} \\
z_{c} \\
1 \\
\end{bmatrix}
\end{aligned}
\end{equation}

After this transformation, point $P_L$ is represented in the LiDAR's frame with $z$ pointing upward. Its projection to the LiDAR's plane thus can be conducted by simply setting its $z$ coordinate to zero. Fig. \ref{fig:projectdata} depicts the projected point cloud.

\subsection{Multilayer fusion map}
\begin{figure}[!ht]
     \centering
     \begin{subfigure}[b]{0.25\textwidth}
         \centering
         \includegraphics[width=\textwidth]{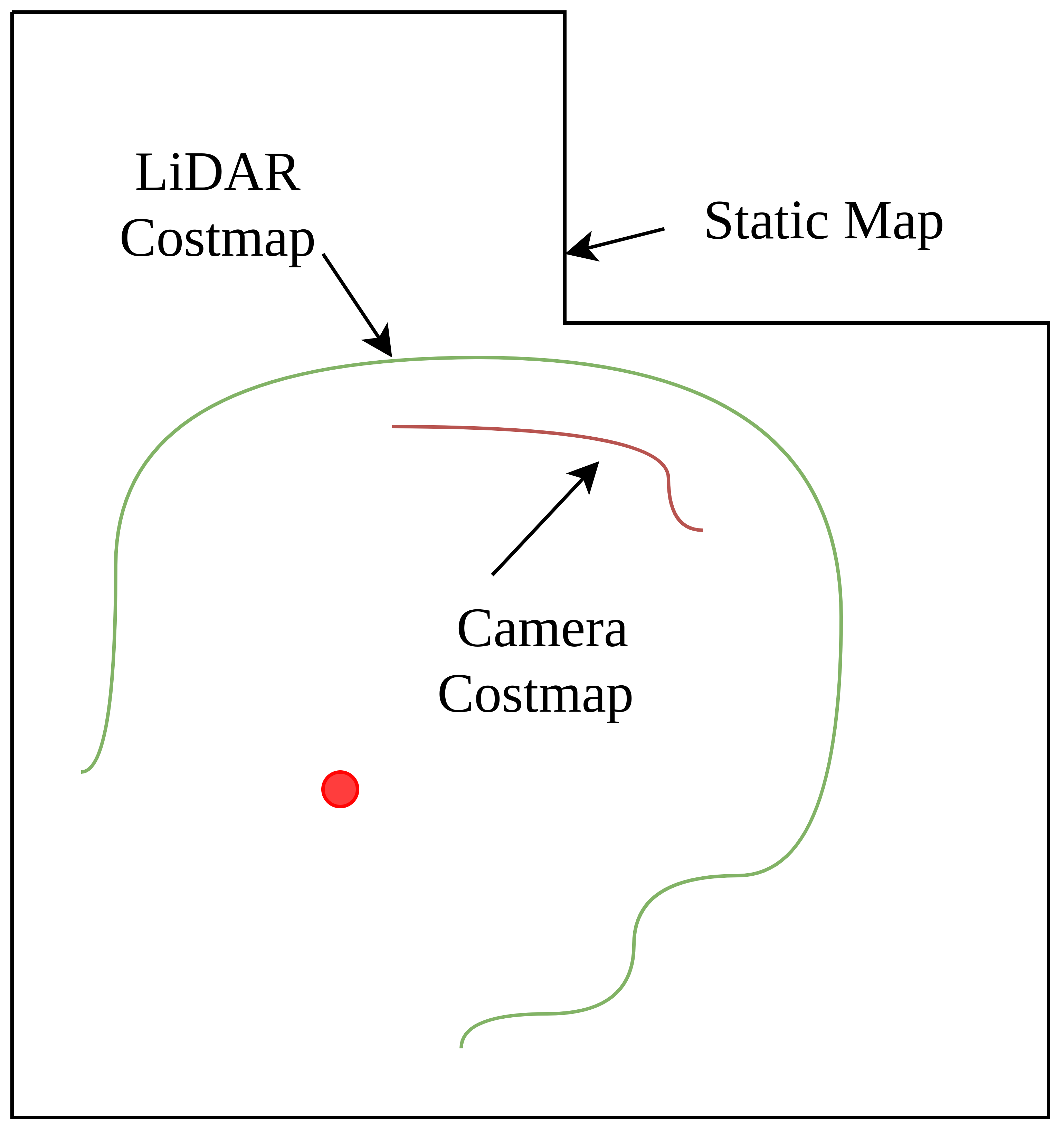}
         \caption{Description of the layers in the map}
         \label{fig:fusionmap1}
     \end{subfigure}
     \hfill
     \begin{subfigure}[b]{0.15\textwidth}
         \centering
         \includegraphics[width=\textwidth]{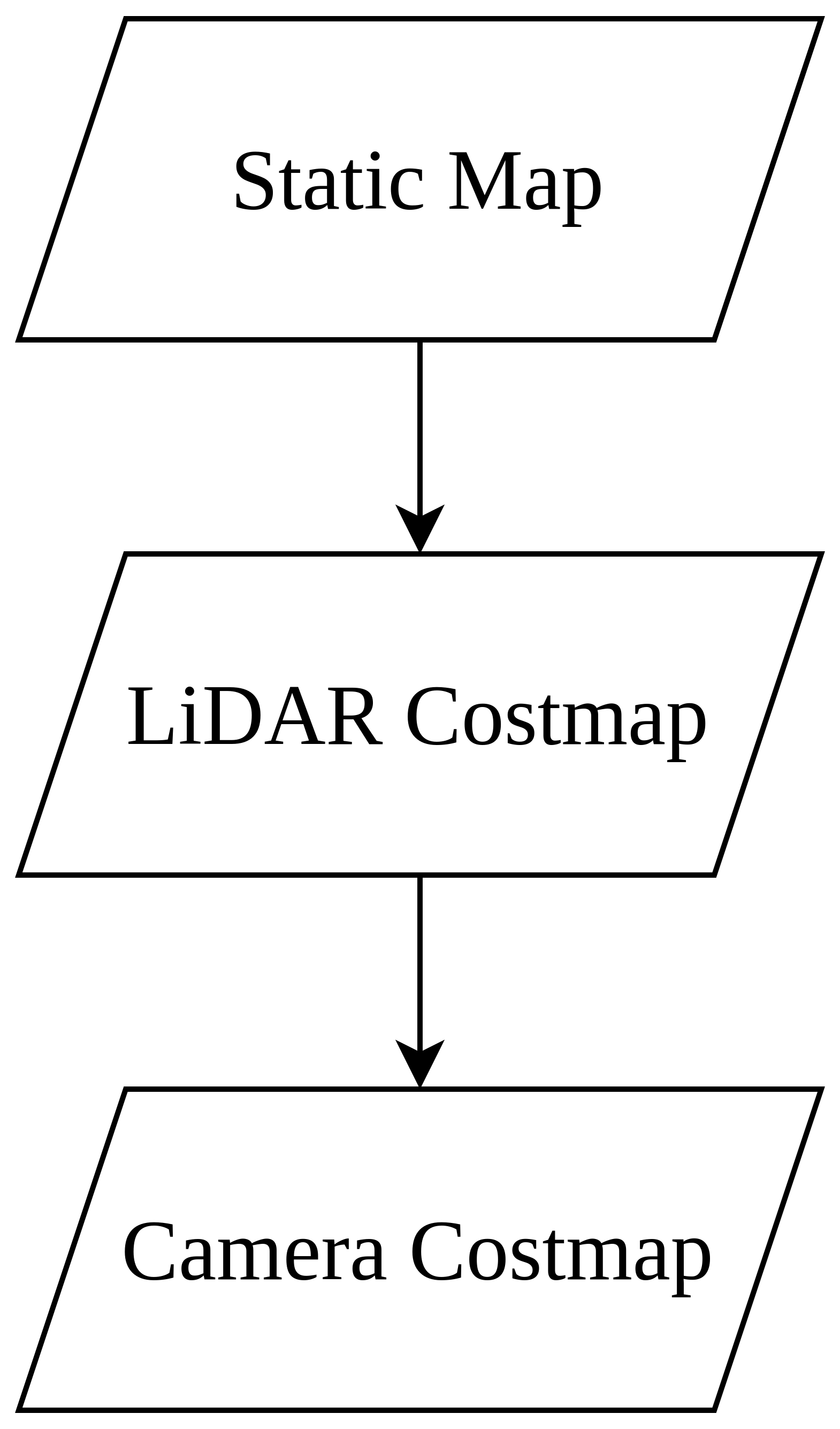}
         \caption{Pipeline of the multi-layer map}
         \label{fig:fusionmap2}
     \end{subfigure}
     \hfill
    \caption{Multilayer fusion map}
    \label{fig:multilayer}
\end{figure}

We present the fused data in a map with three layers named static, LiDAR and camera, as shown Fig. \ref{fig:multilayer}. The static layer represents the map of the whole working area. It is a binary grid map with value ``0" for free cells and ``1" for occupied cells. The LiDAR layer represents the data collected by the LiDAR during robot operation. It is a grid costmap where the value of each cell is ranged from 0 to 255 corresponding to the probability of 0\% to 100\% the cell being occupied. The camera layer is similar to the LiDAR layer except that it is created based on the data collected from the two cameras. During operation, the robot uses information from all three layers to reason, plan and generate control signals. For example, a cell is determined as ``free" only if its value in the static map is ``0" and its values in both the LiDAR and camera costmaps are less than 128. In this way, data from all sensors can be effectively fused to provide reliable information for navigation and obstacle avoidance.

\section{Obstacle avoidance algorithm}\label{AA}
\label{section:ObstacleAvoidance}
Based on the data fused from sensors, we developed an obstacle avoidance algorithm based on the Dynamic Window Approach method with details as follows.
\subsection{Robot kinematics}
The robot used in this work is a differential-drive robot moving on the ground plane so that its pose is determined by its position $(x,y)$ and orientation $\theta$. The kinematic equations of the robot are given by: 

\begin{equation}
\label{Eq:kinematic_Robot}
\begin{aligned}
& \dot{x} = vcos (\theta)\\
& \dot{y} = vsin (\theta)\\
& \dot{\theta} = \omega
\end{aligned}
\end{equation}
where $v$ and $\omega$ are respectively the linear and angular velocities of the robot. Since the robot is controlled by an embedded computer, it is necessary to discretize the kinematic equations. Let $\Delta t$ be the sampling time and $(x_t,y_t,\theta_t)$ be the robot's pose at time step $t$. The discrete kinematic equations are then given by:

\begin{equation}
\begin{aligned}
& x_{t+1} = x_t + v_t \Delta t cos (\theta_t)\\
& y_{t+1} = y_t + v_t \Delta t sin (\theta_t)\\
& \theta_{t+1} = \theta_t + \omega_t \Delta t
\end{aligned}
\label{Eq:kinematicRobot}
\end{equation}
In addition, ranges of the velocities are limited to:
\begin{equation}
\begin{aligned}
\label{Eq:limitvel}
V_m = \{(v, \omega) | v \in [v_{min}, v_{max}], \omega \in [\omega_{min}, \omega_{max}]\},
\end{aligned}
\end{equation}
where $v_{max}$ and $v_{min}$ are the maximum and minimum linear velocities, and $\omega_{max}$ and $\omega_{min}$ are the maximum and minimum angular velocities, respectively. Their linear and angular acceleration and deceleration are limited to $\dot{v}_a, \dot{\omega}_a$ and $\dot{v}_b, \dot{\omega}_b$, respectively.

\subsection{Dynamic window approach}
The obstacle avoidance algorithm used in this work is developed based on the dynamic window approach (DWA) \cite{Fox1997}. This algorithm aims to find a pair of linear and angular velocity values that describe the best trajectory that the robot can achieve within its next time step. Those values are obtained via two main steps: 
\begin{enumerate}
\item Determining feasible velocities based on constraints on acceleration/deceleration and obstacle avoidance.
\item Choosing the velocities that maximize an objective function.
\end{enumerate}

 In particular, the velocities that the robot can reach within its next time interval are constrained by its acceleration and deceleration limits, i.e.,
\begin{equation}
\begin{aligned}
\label{Eq:V_dDWA}
V_d = \biggl\{ (v, \omega) \bigg|
\begin{array}{*{20}{c}}
v \in [v_c - \dot{v}_b \Delta t, v_c + \dot{v}_a \Delta t], \\
\omega \in [\omega_c - \dot{\omega}_b \Delta t, \omega_c + \dot{\omega}_a \Delta t]
\end{array}
\biggl\}
\end{aligned}
\end{equation}
where $v_c$ and $\omega_c$  are respectively the present linear and angular velocities of the robot. Besides, the velocities also need to ensure that the robot does not reach its closest obstacle. Let $dis(v,\omega)$ be the distance from the robot to its closest obstacle on the trajectory generated by $(v,\omega)$. The velocities that avoid collision are:
\begin{equation}
\begin{aligned}
\label{Eq:V_aDWA}
V_a = \biggl\{ (v, \omega) \bigg|
\begin{array}{*{20}{c}}
v \leq \sqrt{2 \cdot dis(v, \omega) \cdot \dot{v}_b}, \\
\omega \leq \sqrt{2 \cdot dis(v, \omega) \cdot \dot{\omega}_b}
\end{array}
\biggl\}
\end{aligned}
\end{equation}
From (\ref{Eq:limitvel}), (\ref{Eq:V_dDWA}), and (\ref{Eq:V_aDWA}), we can limit the range of velocities to
\begin{equation}
\begin{aligned}
\label{Eq:VDWA}
V = V_m \cap V_d \cap V_a
\end{aligned}
\end{equation}

In the second step, specific values of $(v,\omega)$ are determined from $V$ so that they maximize the following objective function \cite{Fox1997}:
\begin{equation}
\begin{aligned}
\label{Eq:Objectfunction}
G(v, \omega) = \alpha \cdot heading(v, \omega) + \beta \cdot dist(v, \omega) + \gamma \cdot vel(v, \omega),
\end{aligned}
\end{equation}
where $\alpha, \beta, \gamma$ are positive weight coefficients. Function $heading(v, \omega)$ is defined as:
\begin{equation}
\begin{aligned}
\label{Eq:headingDWA}
heading(v, \omega) = 1 -  \frac{\theta}{\pi},
\end{aligned}
\end{equation}
where $\theta$ represents the angle between the robot's heading direction and the target. This function thus measures the angle deviation between the end direction of the robot's trajectory and the target. It is maximal if the robot moves directly towards the target.

Function $dist(v, \omega)$ represents the distance from the trajectory to the closest obstacle. Let $d_o$ be the distance to the closest obstacle on the trajectory and $r_e$ be the obstacles’ expanded radius. We have
\begin{equation}
\begin{aligned}
\label{Eq:distantDWA}
 dist(v, \omega) = 
\begin{cases}
\frac{1}{r} d_o & \text{if $d_o < r$} \\
1 & \text{otherwise}
\end{cases} 
\end{aligned}
\end{equation}

Function $vel(v, \omega)$ simply represents the linear velocity of the robot which is preferably to be fast. It is defined as:
\begin{equation}
\begin{aligned}
\label{Eq:velDWA}
vel(v, \omega) = \frac{v}{v_{max}}
\end{aligned}
\end{equation}

By solving (\ref{Eq:Objectfunction}), the velocities $(v, \omega)$ that maximize $G(v, \omega)$ are the optimal velocities being applied to the robot.

\section{Results}
\label{section:experiment}
To evaluate the performance of the proposed approach, we have conducted a number of experiments with details as follows.

\subsection{Experimental setup}
\begin{figure}[!ht]
    \centering
    \begin{subfigure}[b]{0.45\textwidth}
    \centering
    \includegraphics[width=\textwidth]{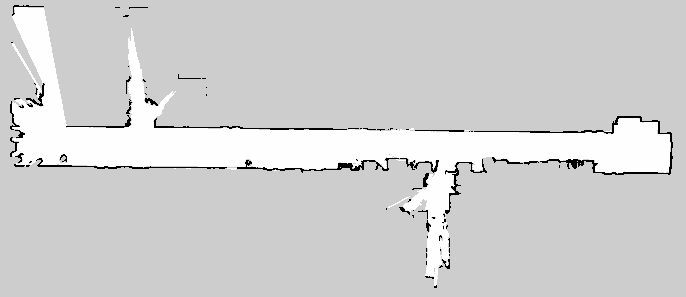}
    \caption{Small corridor}
    \label{fig:env1}
    \end{subfigure}
     \hfill   
    \begin{subfigure}[b]{0.245\textwidth}
    \centering
    \includegraphics[width=\textwidth]{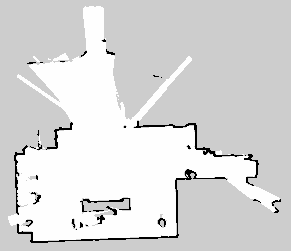}
    \caption{Large room}
    \label{fig:env2}
    \end{subfigure}
    \caption{Two indoor environments used in experiments}
    \label{fig:environment}
\end{figure}

Experiments have been conducted in two environments including a small corridor with the size of 50m $\times$ 4m and a large room with the size of 15m $\times$ 10m as shown in Fig. \ref{fig:environment}.

In experiments, the robot starts from its initial position A to go to its goal B. For the corridor environment, two cases with an AB distance of 10 m and 15 m are evaluated. For the room environment, two cases with the AB distance of 5 m and 10 m have been experimented. During experiments, the Cartographer SLAM algorithm was used to build maps of the environments and localize the robot. A Marvel Mind indoor locator is used to accurately estimate the robot's position, which is then used as the ground truth for comparison. 

\subsection{Experimental results}
\begin{figure}[!ht]
    \centering
    \begin{subfigure}[b]{0.22\textwidth}
    \centering
    \includegraphics[width=\textwidth]{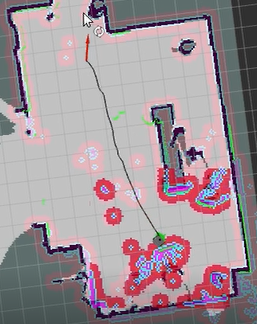}
    \caption{The robot's trajectory when there are no obstacles}
    \label{fig:originalpath}
    \end{subfigure}
    \begin{subfigure}[b]{0.22\textwidth}
    \centering
    \includegraphics[width=\textwidth]{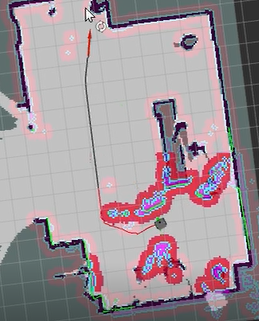}
    \caption{The robot's trajectory when there are obstacles}
    \label{fig:newpath}
    \end{subfigure}
    
    \caption{Robot trajectories in the room environment}
    \label{fig:robot_trajec}
\end{figure}

Figure \ref{fig:robot_trajec} shows experimental results for the room environment where the black lines represent the static map, the green areas represent the LiDAR costmap, and the red areas represents obstacles. It can be seen that, without obstacles, the robot moves straight toward the goal (Fig. \ref{fig:originalpath}). When obstacles appear, the robot changes its trajectory to avoid them and then navigates to the goal as shown in Fig. \ref{fig:newpath}.

\begin{figure}[!ht]
    \centering
    \begin{subfigure}[b]{0.2\textwidth}
    \centering
    \includegraphics[width=\textwidth]{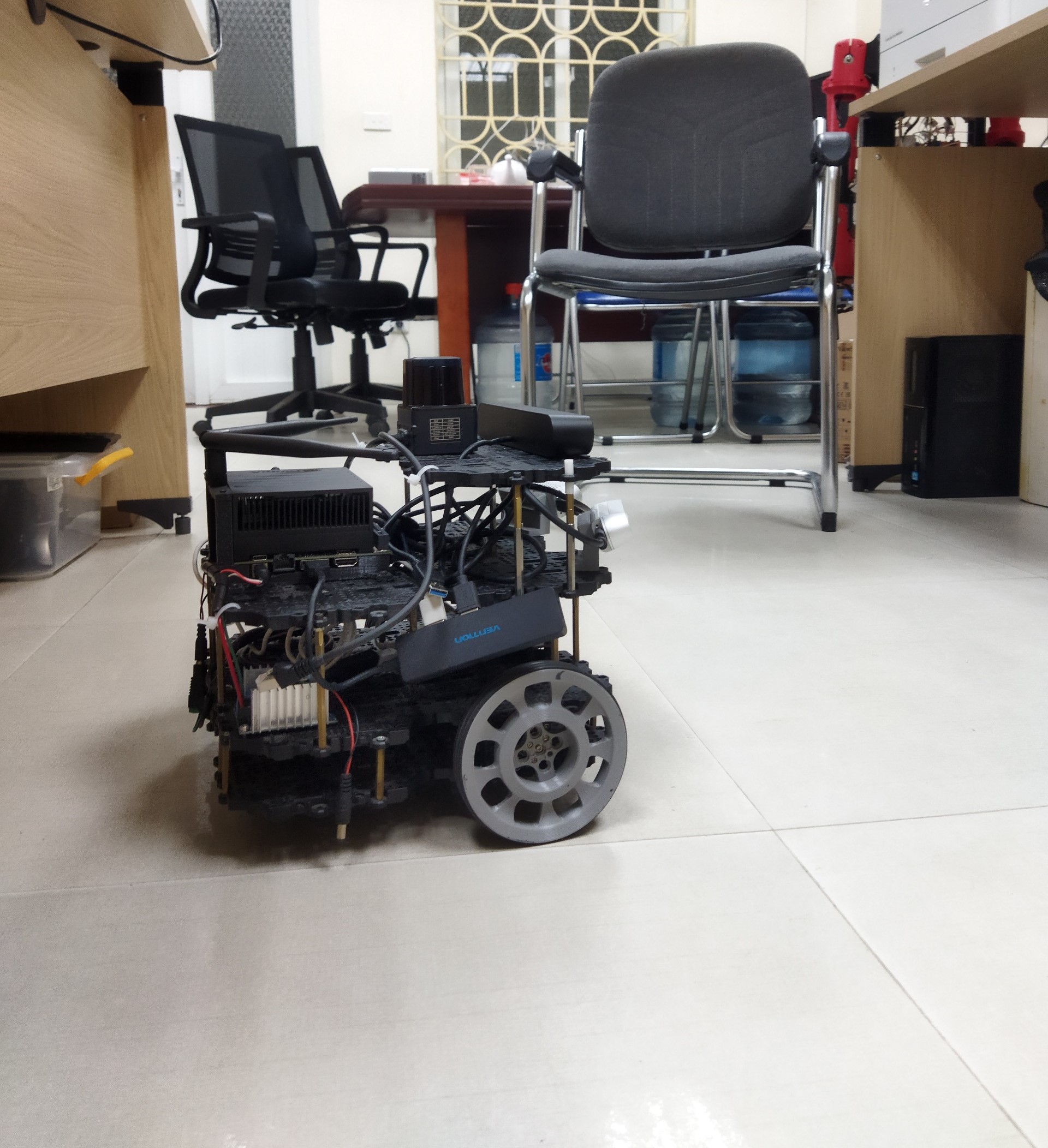}
    \caption{Real world}
    \label{fig:real_world}
    \end{subfigure} \\
    \centering
    \begin{subfigure}[b]{0.2\textwidth}
    \centering
    \includegraphics[width=\textwidth]{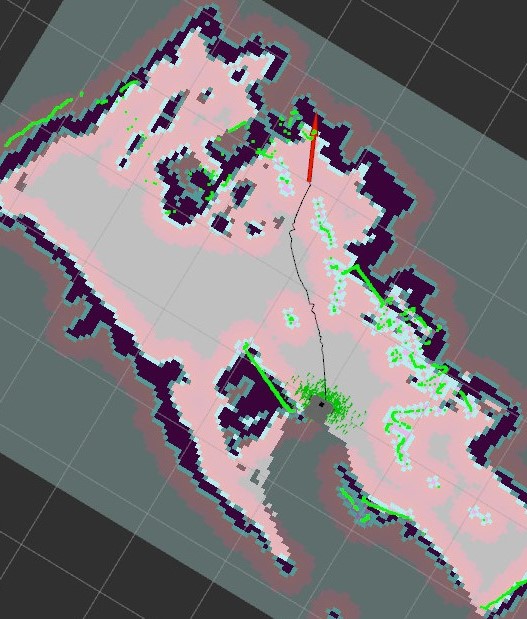}
    \caption{Obstacle avoidance path using only the LiDAR sensor}
    \label{fig:fusion_oldpath}
    \end{subfigure}
    \centering
    \begin{subfigure}[b]{0.2\textwidth}
    \centering
    \includegraphics[width=\textwidth]{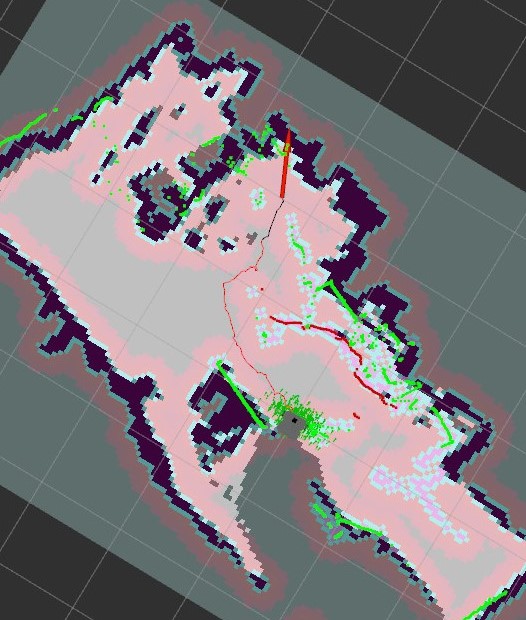}
    \caption{Obstacle avoidance path using sensor fusion \vspace{0.32cm}}
    \label{fig:fusion_newpath}
    \end{subfigure}
    
    \caption{Avoiding chair obstacles in the room environment}
    \label{fig:fusion_robot_trajec}
\end{figure}

To further evaluate the effectiveness of the fusion map, we test the algorithm with chair obstacles as shown in Fig. \ref{fig:real_world}. These are challenging obstacles for the LiDAR since it can only scan the legs of the chair which often leads to wrong reasoning as shown in Fig. \ref{fig:fusion_oldpath}. However, when combined with the camera costmap, the robot can properly detect the obstacles to avoid and navigate to the goal as shown in Fig.\ref{fig:fusion_newpath}.

\begin{figure}[!ht]
    \centering
    \begin{subfigure}[b]{0.25\textwidth}
    \centering
    \includegraphics[width=\textwidth]{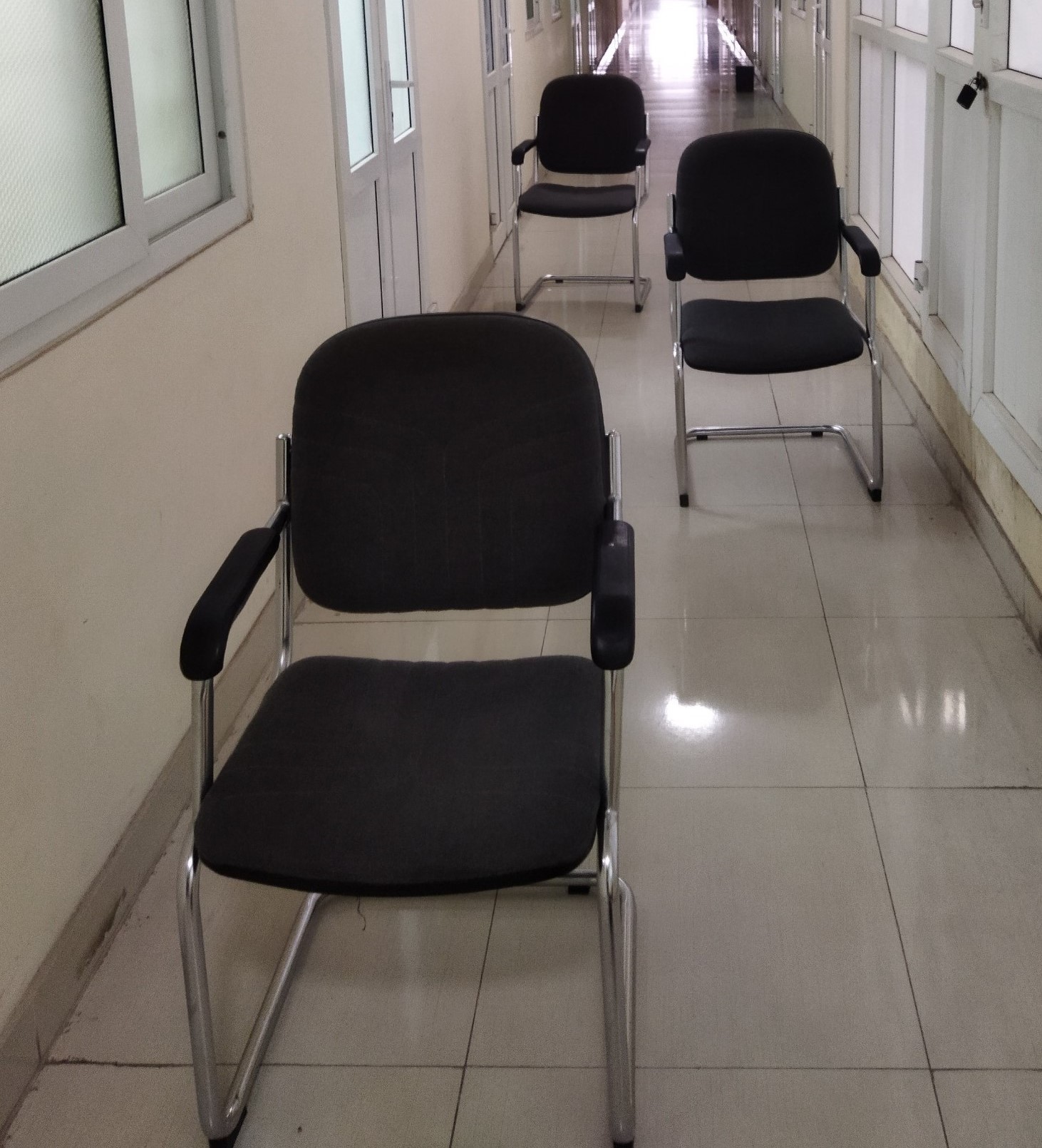}
    \caption{Real world}
    \label{fig:real_chair}
    \end{subfigure}
    \begin{subfigure}[b]{0.2\textwidth}
    \centering
    \includegraphics[width=0.7\textwidth]{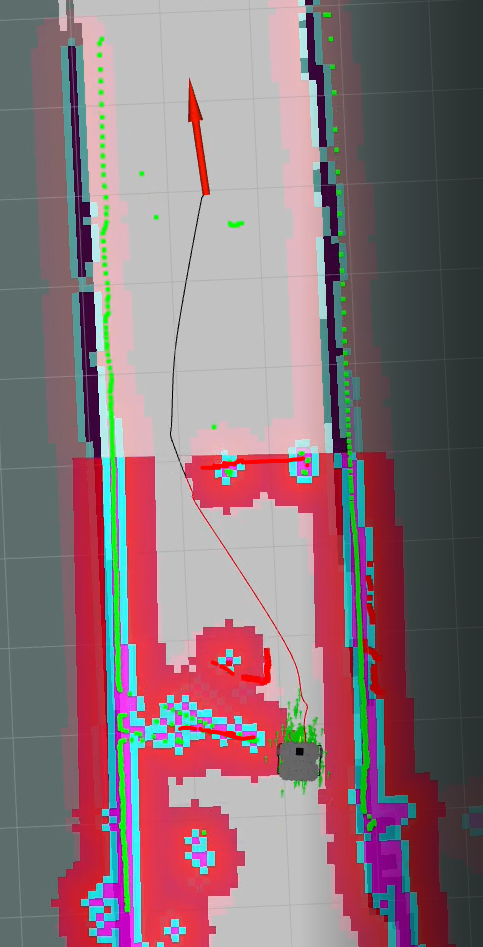}
    \caption{Obstacle avoidance path}
    \label{fig:AvoidPath_Chair}
    \end{subfigure}
    
    \caption{Avoiding chair obstacles in the corridor environment}
    \label{fig:avoid_chair}
\end{figure}

\begin{figure}[!ht]
    \centering
    \begin{subfigure}[b]{0.24\textwidth}
    \centering
    \includegraphics[width=\textwidth]{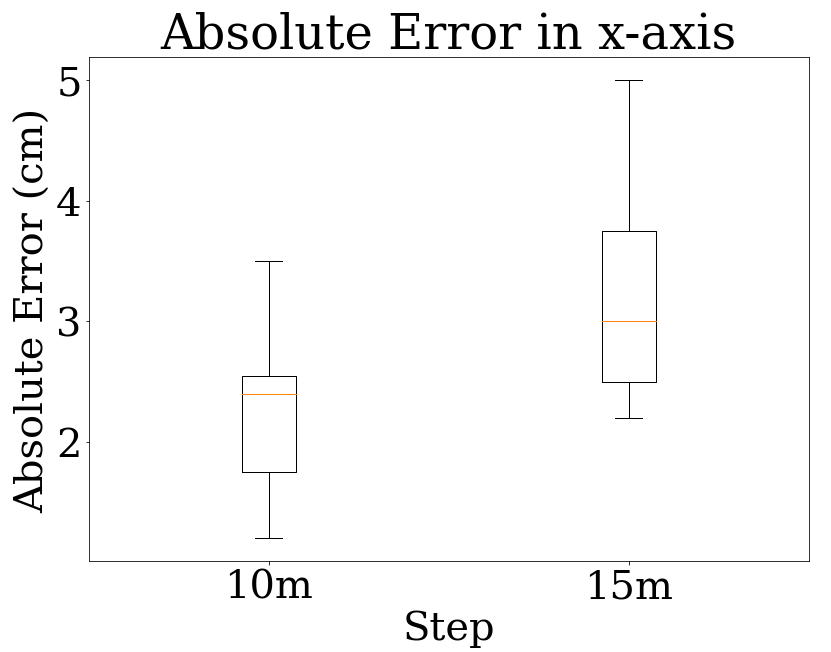}
    \caption{}
    \label{fig:errorX_env1}
    \end{subfigure}
    \begin{subfigure}[b]{0.24\textwidth}
    \centering
    \includegraphics[width=\textwidth]{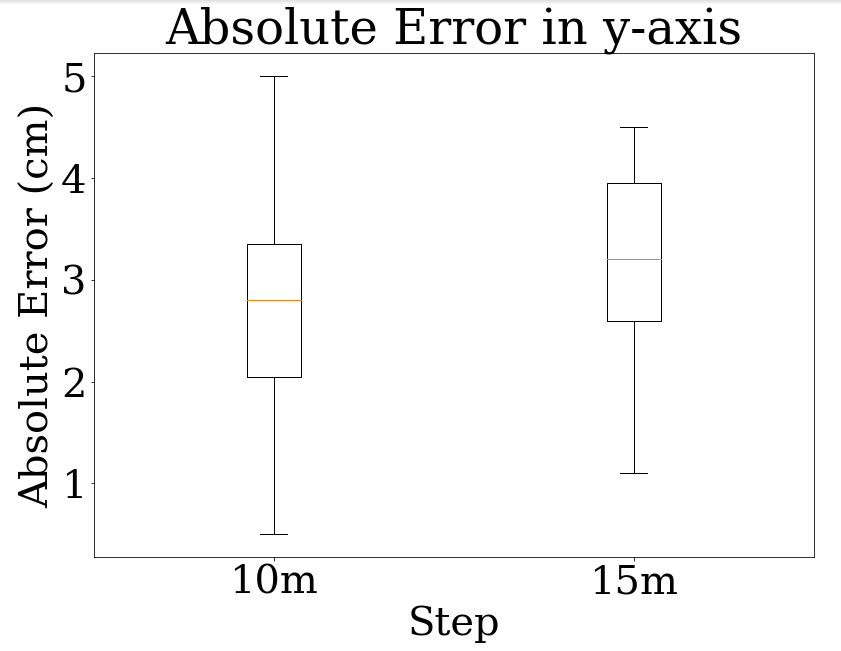}
    \caption{}
    \label{fig:errorY_env1}
    \end{subfigure}
    \caption{Positioning errors during navigation in the corridor environment}
    \label{fig:error_env1}
\end{figure}

\begin{figure}[!ht]
    \centering
    \begin{subfigure}[b]{0.24\textwidth}
    \centering
    \includegraphics[width=\textwidth]{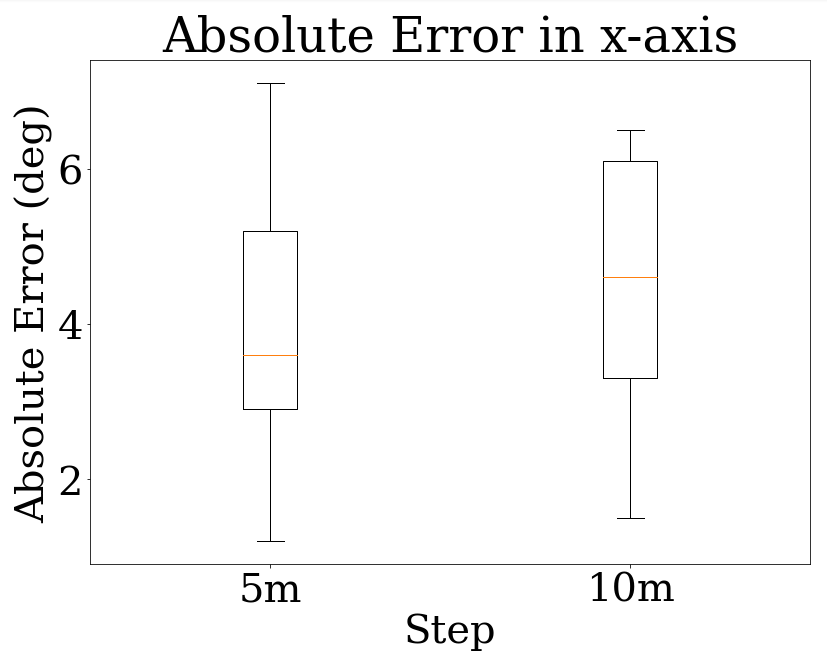}
    \caption{}
    \label{fig:errorX_env2}
    \end{subfigure}
    \begin{subfigure}[b]{0.24\textwidth}
    \centering
    \includegraphics[width=\textwidth]{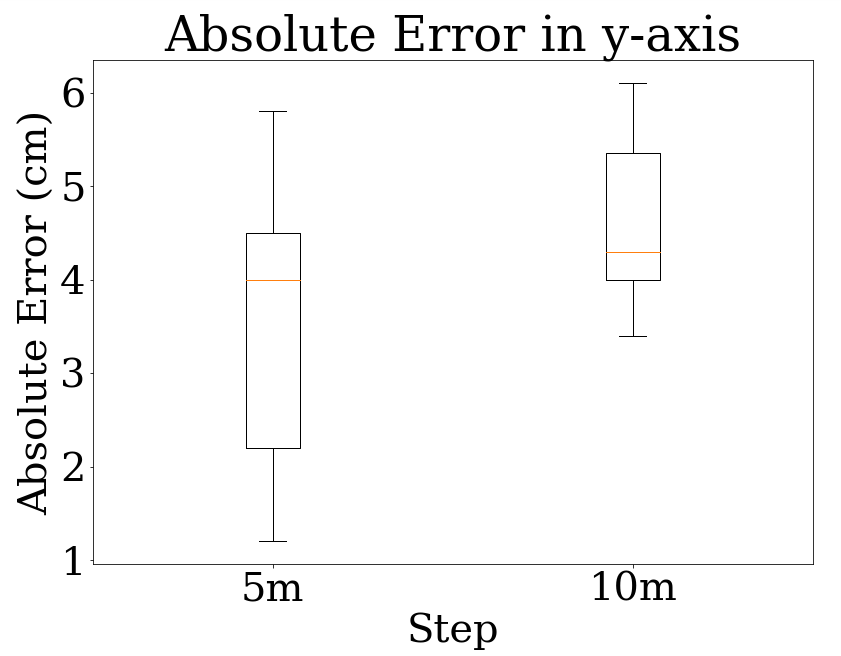}
    \caption{}
    \label{fig:errorY_env2}
    \end{subfigure}
    \caption{Positioning errors during navigation in the room environment}
    \label{fig:error_env2}
\end{figure}

\begin{figure}[!ht]
    \centering
    \begin{subfigure}[b]{0.25\textwidth}
    \centering
    \includegraphics[width=\textwidth]{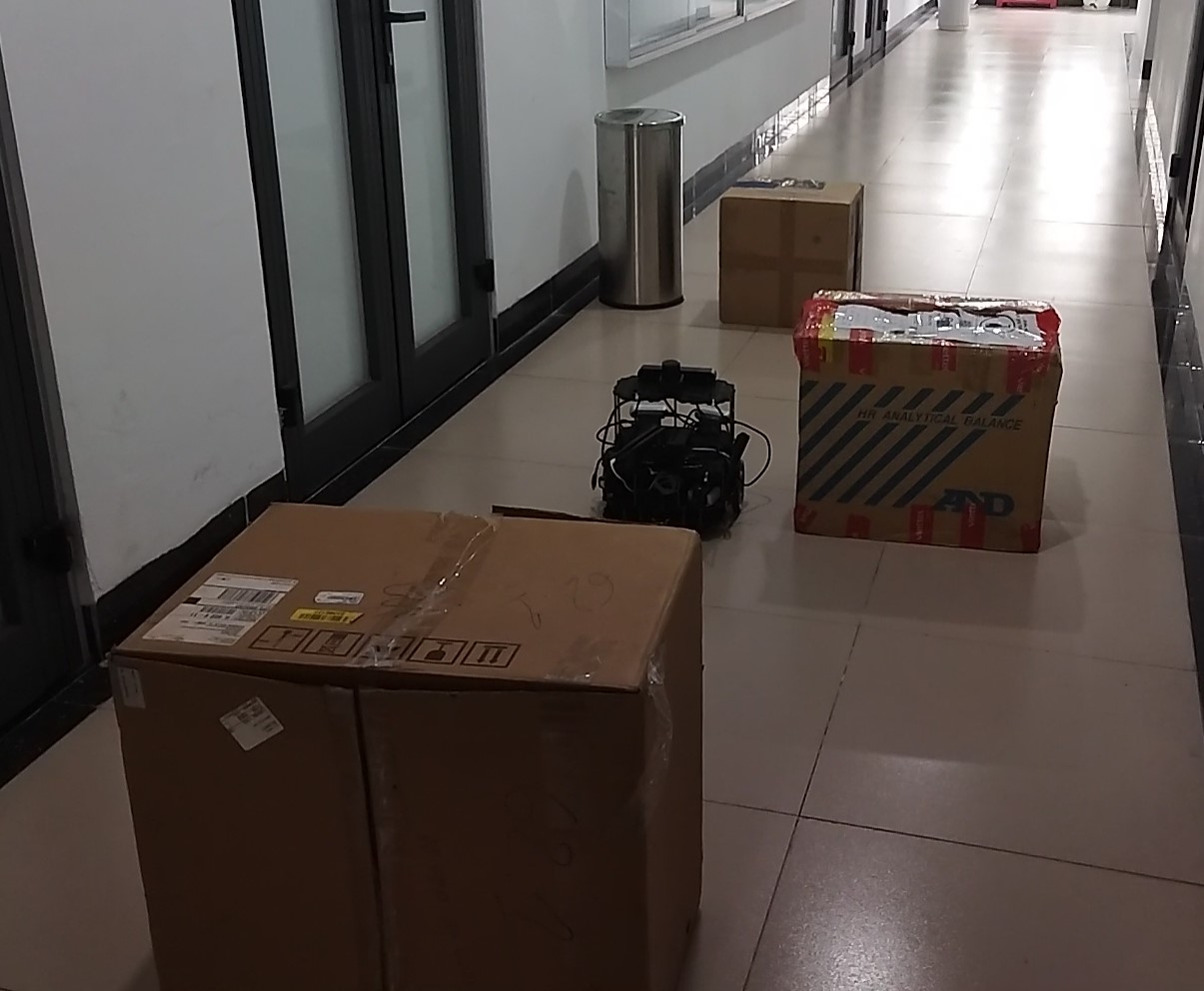}
    \caption{Real world}
    \label{fig:real_static}
    \end{subfigure}
    \begin{subfigure}[b]{0.2\textwidth}
    \centering
    \includegraphics[width=0.75\textwidth]{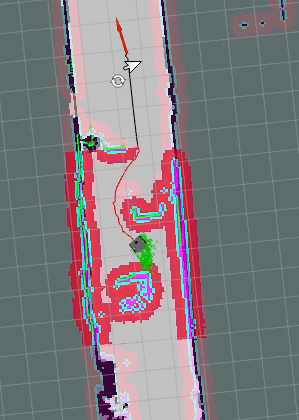}
    \caption{Obstacle avoidance path}
    \label{fig:NewPath_StaticObs}
    \end{subfigure}
    
    \caption{Trajectory of the robot before and after fusion map during navigation in a multi-obstacle environment in a static environment}
    \label{fig:avoid_static_obs}
\end{figure}

For the corridor environment, experiments are also conducted with different types of obstacles including box and chair objects as shown in Fig.\ref{fig:real_chair} and Fig.\ref{fig:real_static}. The results show that the robot can successfully detect those obstacles and avoid them as shown in Fig.\ref{fig:AvoidPath_Chair} and Fig.\ref{fig:NewPath_StaticObs}. These results can be further confirmed via positioning errors as shown in Fig.\ref{fig:error_env1} and Fig.\ref{fig:error_env2}. It can be seen that the average positioning error in all scenarios is around 5 cm which is sufficient for robot operation. 

\begin{figure}[!ht]
    \centering
    \begin{subfigure}[b]{0.2\textwidth}
    \centering
    \includegraphics[width=\textwidth]{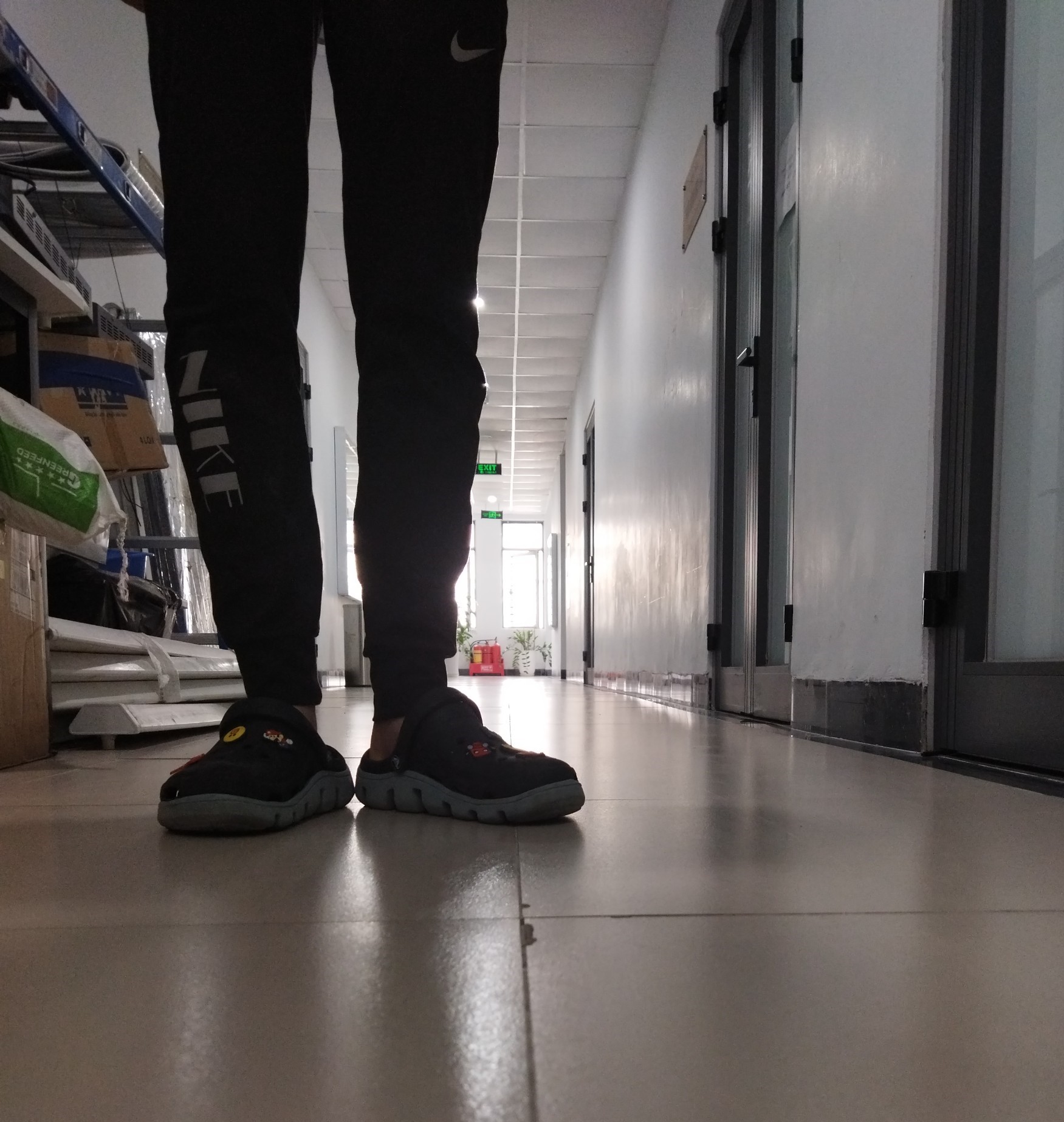}
    \subcaption{Real world \vspace{1.05cm}}
    \label{fig:real_person}
    \end{subfigure}
    \centering
    \begin{subfigure}[b]{0.12\textwidth}
    \centering
    \includegraphics[width=\textwidth]{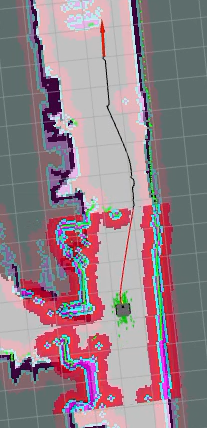}
    \subcaption{The path of the robot when the person is far away}
    \label{fig:Path_DynamicObs}
    \end{subfigure}
    \begin{subfigure}[b]{0.12\textwidth}
    \centering
    \includegraphics[width=\textwidth]{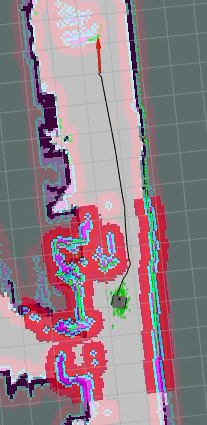}
    \subcaption{The path of the robot when the person is near}
    \label{fig:AvoidPath_DynamicObs}
    \end{subfigure}
    
    \caption{Avoiding dynamic obstacles}
    \label{fig:avoid_person}
\end{figure}

Finally, we evaluate our algorithm in a dynamic environment with a person moving in the corridor as shown in (Fig. \ref{fig:real_person}). Figure \ref{fig:Path_DynamicObs} shows the robot's trajectory when the person is not moving. When he moves, the robot constantly adjusts its trajectory to avoid him as shown in (Fig \ref{fig:AvoidPath_DynamicObs}).

\section{Conclusion}
\label{section:conclusion}
In this paper, we have proposed an approach for reliable obstacle avoidance using the data fused from multiple sensors including depth cameras and LiDAR. To fuse the data, we have introduced a calibration method based on an external camera and a projection technique to convert the 3D data to its 2D correspondence. We have also developed an obstacle avoidance algorithm based on the fused data and DWA. The experimental results show that our algorithm can navigate the robot to avoid different types of static and dynamic obstacles in different environments. The fused data also enhances the positioning accuracy allowing the robot to be applicable in various applications. 

\bibliographystyle{IEEEtran}
\bibliography{ref}
\vspace{12pt}
\color{red}

\end{document}